\documentclass[10pt,twocolumn,letterpaper]{article}

\usepackage{acpr}
\usepackage{times}
\usepackage{epsfig}
\usepackage{graphicx}
\usepackage{amsmath}
\usepackage{amssymb}
\usepackage{graphicx}
\usepackage{subfigure}
\usepackage{multirow}
\usepackage{amsmath}
\usepackage{url}
\usepackage{tikz}
\usepackage{balance}
\usepackage{adjustbox,lipsum}
\usepackage[ruled,vlined]{algorithm2e}

\usepackage[pagebackref=true,breaklinks=true,letterpaper=true,colorlinks,bookmarks=false]{hyperref}

\acprfinalcopy 

\ifacprfinal\fi
\begin{document}
\newcommand*{\textred}{\textcolor{red}}
\newcommand*{\textblue}{\textcolor{blue}}
\newcommand*{\textgreen}{\textcolor{green}}
\newcommand{\repeatthanks}{\textsuperscript{\thefootnote}}
\title{A deep learning framework for segmentation of retinal layers from OCT images}

\author{Karthik Gopinath \thanks{\scriptsize equal contribution} \qquad Samrudhdhi B Rangrej \repeatthanks \qquad Jayanthi Sivaswamy\\
CVIT, IIIT-Hyderabad, India
}

\maketitle

\begin{abstract}
Segmentation of retinal layers from Optical Coherence Tomography (OCT) volumes is a fundamental problem for any computer aided diagnostic algorithm development. This requires preprocessing steps such as denoising, region of interest extraction, flattening and edge detection all of which involve separate parameter tuning.  In this paper, we explore deep learning techniques to automate all these steps and handle the presence/absence of pathologies. A model is proposed consisting of a combination of Convolutional Neural Network (CNN) and Long Short Term Memory (LSTM). The CNN is used to extract layers of interest image and extract the edges, while the LSTM is used to trace the layer boundary. This model is trained on a mixture of normal and AMD cases using minimal data. Validation results on three public datasets show that the pixel-wise mean absolute error obtained with our system is $1.30\pm0.48$ which is lower than the inter-marker error of $1.79\pm0.76$. Our model's performance is also on par with the existing methods. 
\end{abstract}

\section{INTRODUCTION}

Optical Coherence Tomography (OCT) is an imaging modality capable of capturing structural composition of biological tissues at micrometer resolutions. It is popular in ophthalmology for clinical diagnosis of retinal diseases. The structural layers visible in OCT images comprise of seven layers which are (from top to bottom): Retinal Nerve Fiber Layer (RNFL), Ganglion Cell Layer and Inner Plexiform Layer (GCL+IPL), Inner Nuclear Layer (INL), Outer Plexiform Layer (OPL), Outer Nuclear Layer (ONL), Inner Segment (IS), Outer Segment (OS), Retinal Pigment Epithelium (RPE). The boundaries between these layers are of interest for diagnosing retinal diseases like Age related Macular Degeneration, Glaucoma and Cystoidal Macular Edima. Assessment of these diseases demands accurate layer markings. Manual marking of the layers is laborious and time consuming.  

Automating the layer segmentation task is challenging due to the presence of speckle noise, vessel shadows and varying layer orientation. These were generally handled using a set of modules during preprocessing. Denoising relied on methods such as median filtering, block-matching and 3D filtering \cite{dabov2007image} and diffusion filtering \cite{gilboa2004image} \cite{fu2016retinal} \cite{fernandez2005automated}. Vessel shadows were removed by explicitly masking out such regions found via vertical projections \cite{lu2010automated}. Variable orientation of the layers across the dataset was  addressed by flattening the structures with respect to one of the roughly estimated layers \cite{chiu2010automatic}. All these steps are data dependent and hence require tuning. In layer segmentation, the uppermost and the lowermost boundaries (Vitreous-RNFL and RPE-Choroid) are marked by intensity gradients and hence this information has been used to extract them\cite{tan2008mapping} \cite{fabritius2009automated}. Gradient and intensity information along with an active contour approach \cite{mishra2009intra} has also been proposed. By far the most popular approach is based on graph search.  In this class there are techniques which use intensity, gradient and 3D context based cost function for optimization \cite{garvin2008intraretinal}, shortest path computation with Dijkstra’s algorithm \cite{tian2015real} and graph based tracing of the layer boundary \cite{chiu2010automatic}. These methods detect layers in a sequential fashion by constraining the ROI after each detected layer. Most of these  algorithms were proposed for segmenting retinal layers in a normal case.

Presence of pathologies alter the layer morphology locally  and thus increases the complexity of the problem. Automatic segmentation of 3 layers relevant to age related macular generation and geographic atrophy was proposed \cite{chiu2012validated} by adapting the edge weights used in graph theory and dynamic programming based framework \cite{chiu2010automatic}. More recently, information such as slope similarity and non-associativity of layers as edge weight have been explored to handle pathologies \cite{hussain2016automatic}.

\begin{figure*}[thbp]
	\centering
    	\includegraphics[width=0.9\textwidth]{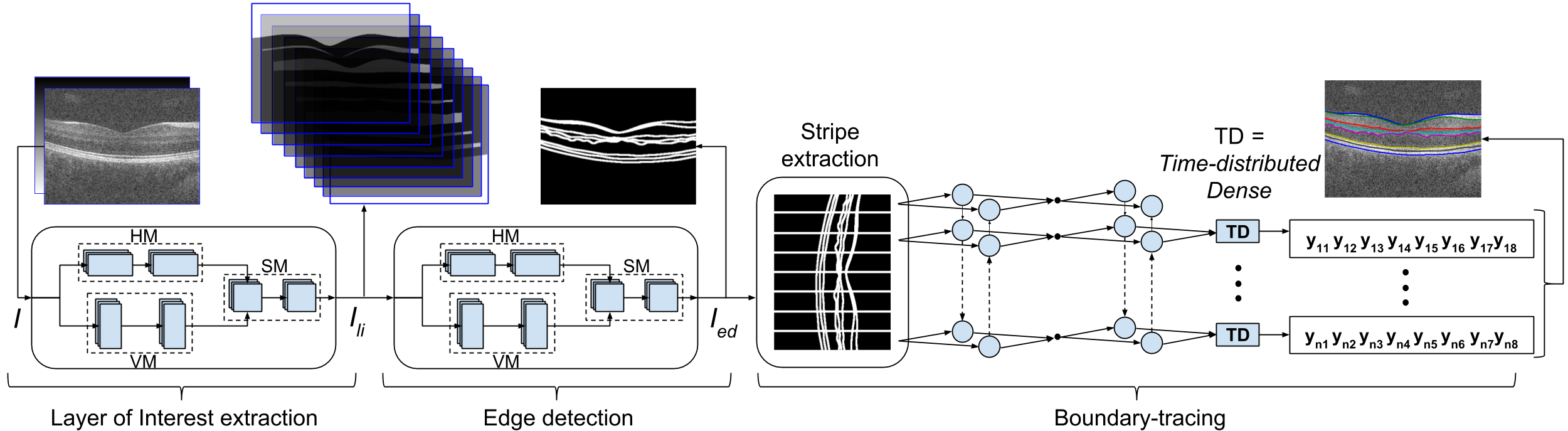}
      \caption{Deep learning architecture for layer segmentation.}
      \label{fig:arch}
\end{figure*}

\begin{table*}[]
\centering
\caption{Description of the proposed architecture}
\label{table:arch_table}
\resizebox{\textwidth}{!}{%
\begin{tabular}{|c|c|c|c|c|c|c|c|c|c|c|c|c|c|}
\hline
\multicolumn{4}{|c|}{LOI extraction} & \multirow{8}{*}{} & \multicolumn{4}{c|}{Edge detection} & \multirow{8}{*}{} & \multicolumn{4}{c|}{Boundary-tracing} \\ \cline{1-4} \cline{6-9} \cline{11-14} 
Module & Layer & Filter size & ACTN &  & Module & Layer & Filter size & ACTN &  & Module & Layer & \begin{tabular}[c]{@{}c@{}}Output\\ Nodes\end{tabular} & ACTN. \\ \cline{1-4} \cline{6-9} \cline{11-14} 
HM1 & CONV & (20x30x2)x32 & Relu~\cite{relu} &  & HM1 & CONV & (15x20x9)x16 & Relu &  & fwd-LSTM1 & LSTM & 64 & - \\  
HM2 & CONV & (20x30x32)x32 & Relu &  & HM2 & CONV & (15x20x16)x16 & Relu &  & bck-LSTM1 & LSTM & 64 & - \\ 
VM1 & CONV & (30x20x2)x32 & Relu &  & VM1 & CONV & (20x15x9)x16 & Relu &  & fwd-LSTM2 & LSTM & 32 & - \\  
VM1 & CONV & (30x20x32)x32 & Relu &  & VM1 & CONV & (20x15x16)x16 & Relu &  & bck-LSTM2 & LSTM & 32 & - \\ 
SM1 & CONV & (10x10x64)x32 & Relu &  & SM1 & CONV & (10x10x32)x16 & Relu &  & \multirow{2}{*}{TD} & \multirow{2}{*}{\begin{tabular}[c]{@{}c@{}}Fully\\ connected\end{tabular}} & \multirow{2}{*}{8} & \multirow{2}{*}{Sigmoid} \\ 
SM2 & CONV & (5x5x32)x9 & Sigmoid &  & SM2 & CONV & (5x5x16)x1 & Sigmoid &  &  &  &  &  \\ \hline
\end{tabular}%
}
\end{table*}

Thus, existing techniques employ \textit{many} (pre)processing blocks all of which require separate tuning of parameters and modify approaches designed for normal cases to handle pathologies. This limits the robustness of the methods. Deep neural networks offer a way to learn the main segmentation task in addition to these early processes. In this paper, we propose a novel supervised method for layer segmentation applicable to both normal and pathology cases. It is based on a combination of Convolutional Neural Network (CNN) and Bidirectional Long Short-term Memory (BLSTM). The major strengths of the proposed method are (i) no requirement for any preprocessing (ii) multi-layer segmentation in one go (iii) robustness to presence of pathologies (iv) robustness to imaging systems and image quality.

\section{PROPOSED ARCHITECTURE}
\label{PM}
OCT images are corrupted due to presence of speckle noise whose characteristics vary across scanners. In order to extract layers of interest which is agnostic to the source of data (scanner), we use a CNN-based first stage. The presence of vessel-shadows and pathologies cause the boundaries between retinal layers to be discontinuous. Naive edge detection algorithm fails to extract eight \textit{continuous} boundaries shared by seven layers. Hence, layer segmentation from the output of first stage is achieved using a cascade of stages: edge detection followed by boundary-tracing. A CNN-based strategy is adopted for the former while for the latter, a specific type of Recurrent Neural Network, namely LSTM is adopted. The LSTM stage learns to trace eight continuous boundaries by following the detected edges, with continuity ensured by using a bidirectional LSTM (referred as BLSTM). Detailed description of proposed architecture is presented next.

\subsection{Architecture}
The custom-designed architecture for the proposed system is shown in Fig.\ref{fig:arch} and details are provided in Table~\ref{table:arch_table} with CONV being Convolution layer and ACTN as Activation. We describe each stage in the system next.

\textbf{Stage 1: Layer of interest (LOI) extraction.} The input to the CNN is a stack of OCT image $I$ in addition to a position cue image $I_{pc}$ which is defined as $I_{pc}(x,y) = y$.
The input passes through 2 parallel modules: Horizontal-filter Module (HM) and Vertical-filter Module (VM). Resultant activations are stacked and merged using a Square-filter Module (SM) to generate a LOI image $I_{li}$ with 9 channels, each depicting one region(7 layers plus the vitreous cavity and choroid). Ideally, each channel of $I_{li}$ is expected to be a Boolean image with $1$ denoting inside the respective regions and 0 denoting exterior points. Along with learning interested layers, HM and VM learn intra- and inter-layer characteristics respectively. HM should ideally learn to inpaint pathologies and vessel-shadows, based on the horizontal neighborhood belonging to the same layer. VM should learn to differentiate two neighboring layers. SM ensures equal contribution from both horizontal and vertical neighbors. $I_{li}$ is passed to next stage as well as taken as an independent side-output.

\begin{figure*}[t]
\centering
\hfill \subfigure[]
{\includegraphics[width=0.4 \textwidth]{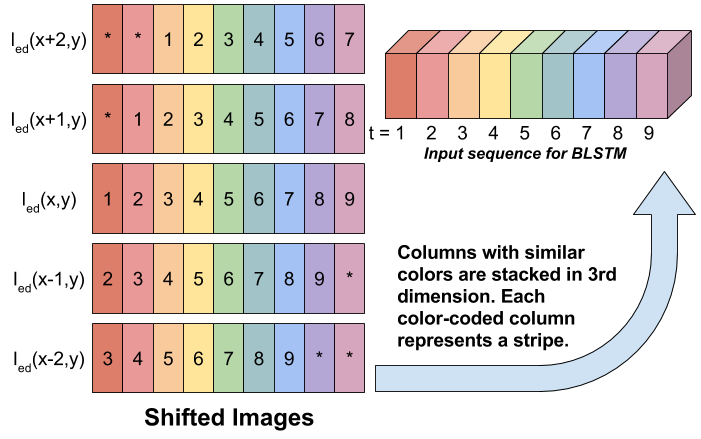}} \hfill \subfigure[]
{\includegraphics[width=0.25 \textwidth]{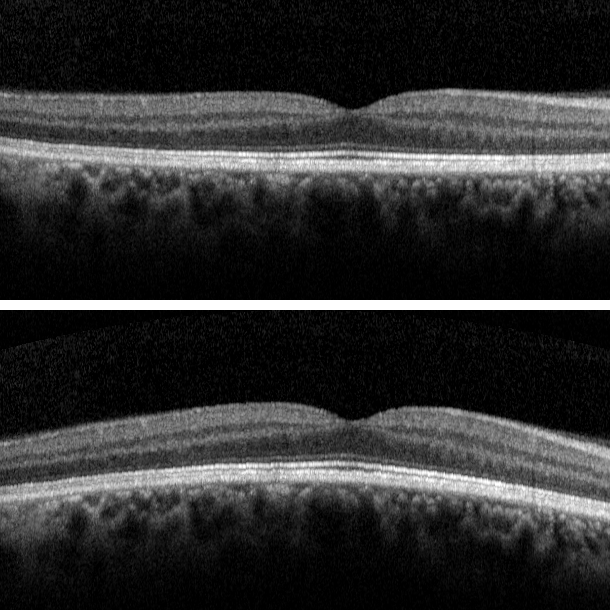}} \hfill \hfill
\caption{(a) Stripe extraction (b) Data augmentation using column rolling. top: original image, bottom: image after column rolling.}
\label{fig:col_roll}
\end{figure*}

\textbf{Stage 2: Edge detection.} This stage is implemented with a second CNN. LOI image $I_{li}$ is passed through HM, VM and SM similar to stage 1. Here HM and VM learn edges with horizontal and vertical orientations respectively. Both edge-maps are combined using SM to generate a single map $I_{ed}$ capturing edges of all orientations. $I_{ed}$ is passed to next stage as well as taken out as a side-output.

\textbf{Stage 3: Boundary-tracing.} This is implemented using a BLSTM. For each column $i$ of an image, eight boundary-coordinates ($L_{i}(j),j \in \{1,2,...,8\}$) depend on the edge passing through the neighboring columns. Hence, a stripe representing information of neighboring columns is extracted (\textit{online}) from $I_{ed}$ as follows. $I_{ed}$ is shifted left and right twice ($I_{ed}(x-k,y); k \in \{0,\pm 1,\pm 2\}$) and stacked such that each column is aligned with its neighbors in the $3^{rd}$ dimension. Each column of this stack is termed as `stripe' (see Fig.\ref{fig:col_roll}(a)). Extracted stripes are sequentially passed to the two-staged BLSTM. A BLSTM has two LSTMs, each learns to generate boundary-coordinates for the current stripe in continuation with coordinates from the right and the left neighboring stripes respectively. Estimated $L_{i},i \in \{1,2,..., No.\;of\;columns\}$ traces the desired 8 layer boundaries simultaneously across the image.


\section{MATRIALS AND TRAINING METHOD}

\subsection{Dataset description}
The proposed system was trained and evaluated using publicly available datasets.

\textbf{Dataset with normal cases:} Two publicly available datasets were considered. The first (we refer to as $Chiu_{norm}$) is made available by Chiu \textit{et al.} \cite{chiu2010automatic} and contains 110 B-scans from 10 healthy subjects (11 B-scans per subject) along with \textit{partial} manual markings from two experts. The authors state that, ``each expert grader exhibited a bias when tracing layer boundaries" and ``manual segmentation tended to be smooth''. Thus, manual segmentation does not follow edges tightly. The second dataset (called OCTRIMA3D) is made available by Tian \textit{et al.} \cite{tian2015real} and contains 100 B-scans from 10 healthy subjects(10 B-scans per subject); manual markings by two observers are also provided. The authors emphasize that in contrast to the smooth manual labellings in \cite{chiu2010automatic}, the delineated boundaries in OCTRIMA3D trace small bumps.

\textbf{Dataset with pathology cases \cite{chiu2012validated}:} This is a dataset made available by Chiu \textit{et al.} \cite{chiu2012validated} and it consists of a total of 220 B-scans from 20 volumes. This dataset is characterized by the presence of pathologies such as drusen and geographic atrophy and hence we refer to it as $Chiu_{path}$. Manual segmentation from two experts are available for only 3 layer boundaries (Vitreous-RNFL, OS-RPE and RPE-Choroid). The dataset includes scans with varying image quality.

\subsection{Preparation of Training data}

\textbf{Input image $I$}. The datasets have been acquired with varying scanning protocols and hence vary in resolution and layer orientation. This is addressed by standardizing the images to $300\times 800$ pixel resolution as follows. Columns of each image were summed to obtain a 1D projection and a Gaussian function was fitted on the same. The mean value of the Gaussian represents the y-coordinate of the center $(C_{ROI}(x,y))$ of a region containing the layers. Images were shifted/cropped/padded vertically such that resultant height is 300px and y-coordinates of $C_{ROI}$ and the center of the image are aligned. Next, images were padded horizontally with trailing zeros to achieve a standardized width of 800px. The position cue image $I_{pc}$ was generated with pixel value proportional to the y-coordinate of the pixel location. Finally, $I$ and $I_{pc}$ were stacked and the pixel intensities were normalized to have zero mean and unit variance. 
 
\textbf{Layer boundary coordinates $L_{i}(j)$}. Since the $Chiu_{norm}$ dataset provides partial/discontinuous manual markings from two experts whereas our network requires unique and continuous GT for each image, partial markings from the $1^{st}$ expert was given to a local expert who completed them. It is to be noted that subjective bias and smoothness of marking maintained by the $1^{st}$ expert is impossible to be reproduced by the local expert. Also, the markers for all three datasets used for training are different and exhibit different subjective biases and smoothness of marking. In summary, GT used for training is noisy given that they are derived from 3 different markers.

\textbf{LOI image $I_{li}$}. Each $I$ can be divided in 9 regions demarcated by 8 boundaries. 9 binary images were defined to represent each region (see Fig \ref{fig:arch}) which were stacked to generate $I_{li}$. 

\begin{figure*}[t]
\centering
\includegraphics[width=0.90 \textwidth]{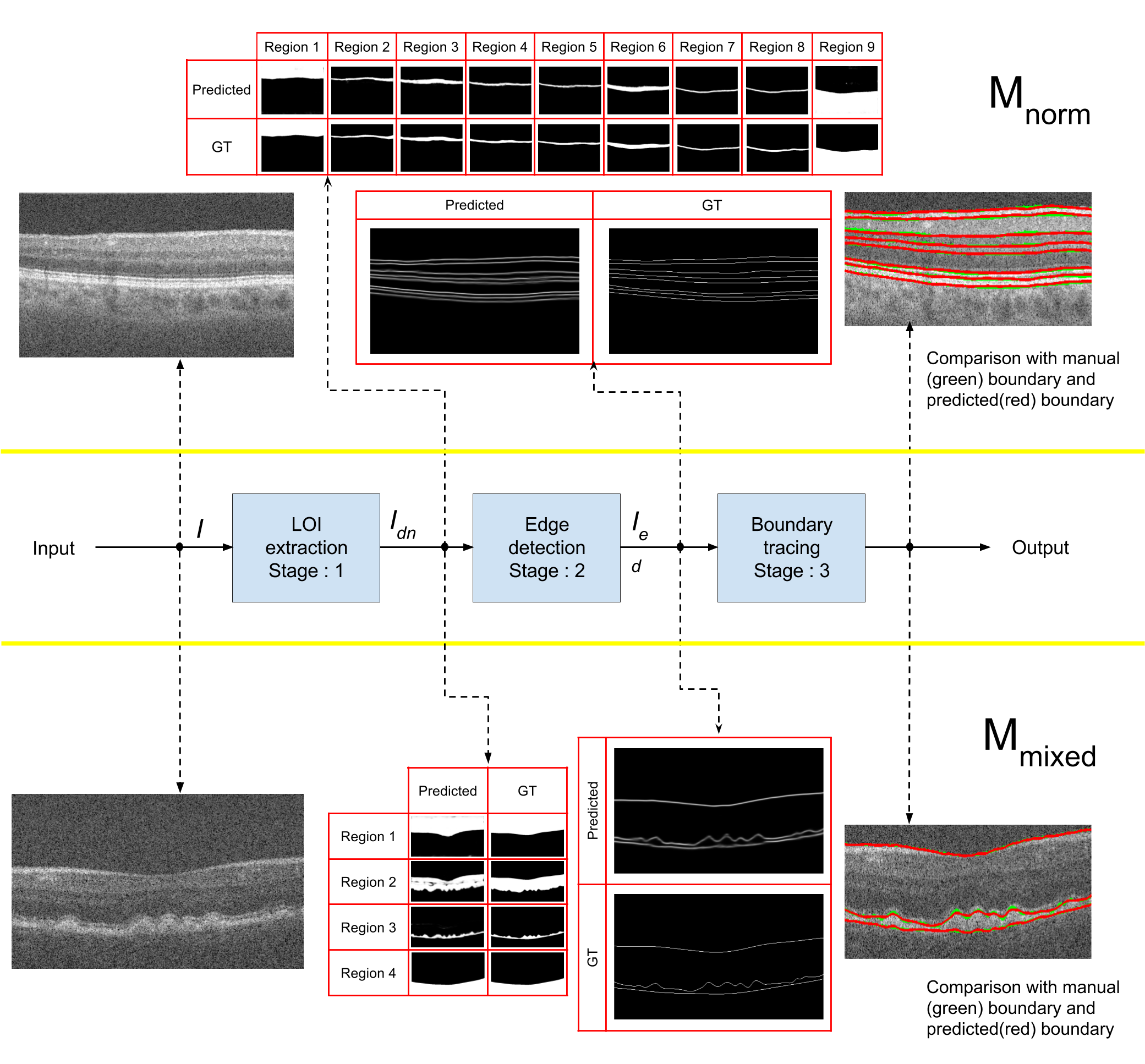}
\caption{Visualization of side output}
\label{fig:side_op}
\end{figure*}

\begin{table*}[t]
\centering
\caption{Pixel-level Mean Absolute Error(MAE) for predictions with $M_{norm}$ network. Values indicate mean$\pm$std.}
\label{table:mae_nor}
\resizebox{\linewidth}{!}{
\begin{tabular}{|l|c|c|c|c|c|c|}
\hline
\multirow{2}{*}{Boundary} & \multicolumn{3}{c|}{$Chiu_{norm}$} & \multicolumn{3}{c|}{OCTRIMA3D} \\ \cline{2-7} 
 & Inter marker & \cite{chiu2010automatic} & ours & Inter marker & \cite{tian2015real} & ours \\ \hline
Vitreous - RNFL & 2.37$\pm$0.79 & 1.38$\pm$0.37 & 1.11$\pm$0.30 & 1.00$\pm$0.24 & 0.68$\pm$0.20 & 1.49$\pm$0.47 \\ 
RNFL - GCL\&IPL & 1.73$\pm$0.85 & 1.67$\pm$0.77 & 1.38$\pm$0.42 & 1.70$\pm$0.76 & 1.16$\pm$0.34 & 1.56$\pm$0.38 \\ 
GCL\&IPL - INL & 1.81$\pm$1.44 & 1.48$\pm$0.58 & 1.42$\pm$0.58 & 1.79$\pm$0.47 & 1.01$\pm$0.15 & 1.24$\pm$0.30 \\ 
INL - OPL & 3.02$\pm$0.87 & 1.48$\pm$0.46 & 1.60$\pm$0.32 & 1.44$\pm$0.33 & 1.11$\pm$0.41 & 1.39$\pm$0.51 \\
OPL - ONL\&IS & 2.18$\pm$0.97 & 1.74$\pm$0.65 & 1.88$\pm$0.65 & 1.83$\pm$0.60 & 1.50$\pm$0.77 & 1.78$\pm$0.77 \\
ONL\&IS - OS & 2.85$\pm$0.93 & 1.00$\pm$0.30 & 0.92$\pm$0.34 & 0.76$\pm$0.22 & 0.54$\pm$0.10 & 0.91$\pm$0.25 \\
OS - RPE & 1.88$\pm$1.08 & 1.14$\pm$0.40 & 1.01$\pm$0.23 & 1.81$\pm$0.87 & 1.22$\pm$0.53 & 1.09$\pm$0.32 \\ 
RPE - Choroid & 2.18$\pm$1.69 & 1.26$\pm$0.35 & 1.43$\pm$0.68 & 1.22$\pm$0.22 & 0.76$\pm$0.17 & 0.98$\pm$0.27 \\ \hline
Overall & 2.25$\pm$1.08 & 1.39$\pm$0.48 & 1.34$\pm$0.44 & 1.44$\pm$0.47 & 1.00$\pm$0.33 & 1.30$\pm$0.41 \\  \hline
\end{tabular}
}
\end{table*}

\begin{table*}[t]
\centering
\caption{Pixel-level Mean Absolute Error for predictions with $M_{mixed}$ network. Values indicate mean$\pm$std.}
\label{table:mae_amd}
\resizebox{\linewidth}{!}{%
\begin{tabular}{|l|c|c|c|c|c|c|c|c|c|}
\hline
\multirow{2}{*}{Boundary} & \multicolumn{3}{c|}{$Chiu_{path}$} & \multicolumn{3}{c|}{$Chiu_{norm}$} & \multicolumn{3}{c|}{OCTRIMA3D} \\ \cline{2-10} 
 & Inter marker & \cite{chiu2012validated} & ours & Inter marker & \cite{chiu2010automatic} & ours & Inter marker & \cite{tian2015real} & ours \\ \hline
Vitreous - RNFL & 1.25$\pm$0.39 & 1.14$\pm$0.339 & 0.95$\pm$0.28 & 2.37$\pm$0.79 & 1.15$\pm$0.32 & 1.02$\pm$0.28 & 1.00$\pm$0.24 & 0.70$\pm$0.25 & 1.03$\pm$0.35 \\
OS - RPE & 2.56$\pm$0.75 & 2.53$\pm$0.83 & 2.41$\pm$0.77 & 1.88$\pm$1.08 & 0.99$\pm$0.20 & 1.12$\pm$0.44 & 1.81$\pm$0.87 & 1.24$\pm$0.58 & 0.87$\pm$0.20 \\ 
RPE - Choroid & 1.55$\pm$0.71 & 1.46$\pm$1.13 & 1.97$\pm$1.23 & 2.18$\pm$1.69 & 1.41$\pm$0.65 & 1.32$\pm$0.67 & 1.22$\pm$0.22 & 0.73$\pm$0.18 & 0.92$\pm$0.33 \\ \hline
Overall & 1.79$\pm$0.62 & 1.71$\pm$0.76 & 1.78$\pm$0.78 & 2.14$\pm$1.19 & 1.18$\pm$0.39 & 1.16$\pm$0.46 & 1.34$\pm$0.45 & 0.89$\pm$0.348 & 0.92$\pm$0.31 \\ \hline
\end{tabular}
}
\end{table*}

\begin{figure*}[thpb]
  \centering
{\includegraphics[width=0.45 \textwidth]{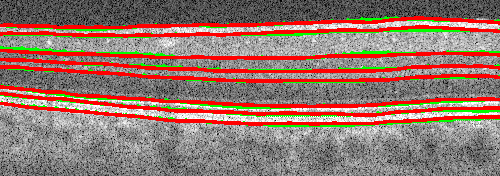}}\quad
  {\includegraphics[width=0.45 \textwidth]{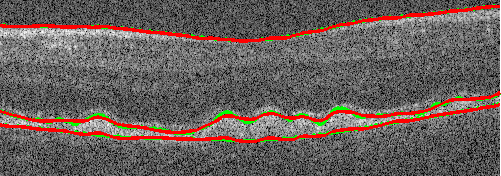}}\quad
  \subfigure[]
  {\includegraphics[width=0.45 \textwidth]{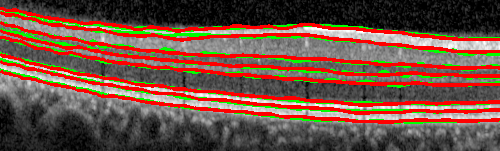}}\quad
  \subfigure[]
  {\includegraphics[width=0.45 \textwidth]{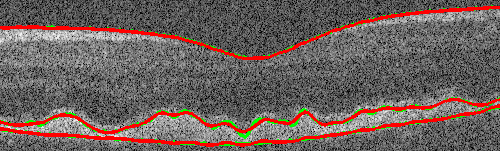}}\quad
  
  \caption{ Comparison with manual (green) segmentation. Results (in red) for a Normal case (a) and a pathological case (b) predicted by $M_{norm}$ and $M_{mixed}$, respectively. }
\label{qual_res}
\end{figure*}

\textbf{Edge-map $I_{ed}$}. For all $I$s, $I_{ed}$ is defined as a binary image indicating $1$ at the boundary pixel and $0$ elsewhere (see Fig \ref{fig:arch}).

As mentioned earlier, GT for pathological cases have markings for only 3 boundaries. Hence, $L_{i}(j)$, $I_{li}$ and $I_{ed}$ for the $Chiu_{path}$ dataset were modified to have only 3 boundaries.

\subsection{Training}
Two copies of the proposed system were trained, one for only normal cases and one for both normal and pathological cases. The first copy, referred to as $M_{norm}$, was trained for 8 layer-boundaries using two normal datasets. The second copy, referred to as $M_{mixed}$ was trained for detecting the 3 boundaries for which manual markings were available. $M_{mixed}$ was trained using the GT for 3 boundaries in $Chiu_{path}$ dataset and the same ones in $Chiu_{norm}$ and OCTRIMA3D. 

For both $M_{norm}$ and $M_{mixed}$, datasets were divided into training and testing set using a split of 8:2. Equal portion of pathological and normal cases were taken for training and testing $M_{mixed}$.  Online data augmentation was done by applying random rotation, scaling, horizontal and vertical flips, shift and column rolling wherein neighboring columns are rolled up/down systematically (see Fig.\ref{fig:col_roll}(b)).Training of the entire network was done in an end-to-end fashion. All slices(11 B-scans) of a single volume constituted a batch while training. Binary cross-entropy and Mean Square Error loss functions were used at stages-1,2 and stage-3 respectively. ADADELTA optimizer \cite{zeiler2012adadelta} was used with sample \textit{emphasizing scheme} \cite{lecun2012efficient}. \textit{Emphasizing scheme} chooses training example with large errors more frequently. This ensures that network sees informative data more often.

The proposed pipeline was implemented on a NVIDIA GTX TITAN GPU, with 12GB of GPU RAM on a core i7 processor. The entire architecture was implemented in Theano using the Keras library. Training for 250 epochs took about a week. The testing time for each OCT volume (11 slices) is 4s.

  


\section{RESULTS}

The proposed system is shown Fig. \ref{fig:side_op} along with a sample input and the outputs of various stages. Both predicted and GT are shown for each stage for the $M_{normal}$ ($M_{mixed}$) in the top (bottom) panel. The model has two side outputs at stages 1 and 2 respectively. The output of the first stage is a 9-channel image, with each channel representing a region (Vitreous, RNFL, GCL+IPL, INL, OPL, ONL+IS, OS, RPE, Choroid). The output of second stage is a single channel image representing the edges between the two consecutive layers.




The networks $M_{norm}$ and $M_{mixed}$ were evaluated separately using the mean absolute pixel error (MAE). $M_{norm}$ was trained for predicting eight layer boundaries on $Chiu_{norm}$ and $OCTRIMA3D$ datasets. $M_{mixed}$ was trained with normal and pathological data ($Chiu_{path}, Chiu_{norm}, OCTRIMA3D$) for estimating three layer boundaries.  Benchmarking for each dataset was done against the output provided by the authors of the datasets. Qualitative results comparing manual marking to predicted layer boundaries are shown in Fig. \ref{qual_res}. Tables \ref{table:mae_nor} and \ref{table:mae_amd} list the obtained MAE for the  $M_{norm}$ and $M_{mixed}$ networks, respectively. The overall MAE obtained by our system is lower than the inter marker error on all the datasets and our results are comparable to the benchmarks at a subpixel level. It is to be noted that MAE for normal cases is less with $M_{mixed}$ than that achieved for the same cases with $M_{norm}$. This is due to the larger training set for $M_{mixed}$ compared to $M_{norm}$. However, the improvement appears to be less for the OS-RPE boundary as the variability to be learnt has also increased. Although training data has increased, it is not enough to learn this increased variability.

\section{CONCLUDING REMARKS} 
\label{sec:diss_con}
We proposed a solution for OCT layer segmentation problem using deep networks. It is in the form of a single system which integrates all the processing steps and segments all the layers in parallel. Such a design obviates the need for tuning individual stages that is commonly found in existing solutions. Results demonstrate that the system is robust to change in scanner and image quality and is able to handle both normal and pathological cases without any data dependent tuning. These are the major strengths of our method. The performance of this system, even with the limited training, is comparable to the existing benchmarks. Increase in the number of training cases led the system to learn more accurate segmentation boundaries. Two notable issues are: i) The ground-truth used for training was sourced from multiple, biased markers resulting in ambiguities in the training process. ii) the precision of the output of the edge-detection stage is at most 1 pixel whereas the BLSTM requires sub-pixel precision. The 1 pixel wide edge causes a confusion field for the BLSTM while tracing precise boundaries. Nevertheless, the proposed system for layer segmentation is robust, fast, automatic with potential for further improvements.

\section*{Acknowledgement}
This work was supported by the Dept. of Electronics and Information Technology, Govt. of India under Grant: DeitY/R\&D/TDC/13(8)/2013.

\balance

{\small
\bibliographystyle{ieee}
\bibliography{egbib}
}

\end{document}